\documentclass[letterpaper, 10 pt, conference]{IEEEtran}  
\IEEEoverridecommandlockouts                              
\usepackage[noadjust]{cite}
\usepackage{graphicx}
\usepackage{amsmath}
\usepackage{algpseudocode}
\usepackage{algorithm}
\usepackage[algo2e]{algorithm2e}
\usepackage{multirow}
\usepackage[normalem]{ulem}
\usepackage{hyperref}

\DeclareMathOperator*{\argmax}{arg\,max}
\hypersetup{
    colorlinks=true,
    linkcolor=blue,
    filecolor=magenta,      
    urlcolor=blue,
}
\usepackage{verbatim}

\newcommand{\app}{\raise.17ex\hbox{$\scriptstyle\sim$}}

\title{\LARGE \bf 
Learning to Grasp without Seeing
}

\author{%
Adithyavairavan Murali, Yin Li, Dhiraj Gandhi, Abhinav Gupta \\
The Robotics Institute, Carnegie Mellon University \\
\texttt{\{amurali, yinl2, dgandhi, gabhinav\}@andrew.cmu.edu}
}%

\begin{document}
\maketitle
\thispagestyle{empty}
\pagestyle{empty}
\begin{abstract}

Can a robot grasp an unknown object without seeing it? In this paper, we present a tactile-sensing based approach to this challenging problem of grasping novel objects without prior knowledge of their location or physical properties. Our key idea is to combine touch based object localization with tactile based re-grasping. To train our learning models, we created a large-scale grasping dataset, including more than $30$K RGB frames and over $2.8$ million tactile samples from $7800$ grasp interactions of $52$ objects. To learn a representation of tactile signals, we propose an unsupervised auto-encoding scheme, which shows a significant improvement of $4$-$9$\% over prior methods on a variety of tactile perception tasks. Our system consists of two steps. First, our touch localization model sequentially ``touch-scans'' the workspace and uses a particle filter to aggregate beliefs from multiple hits of the target. It outputs an estimate of the object's location, from which an initial grasp is established. Next, our re-grasping model learns to progressively improve grasps with tactile feedback based on the learned features. This network learns to estimate grasp stability and predict adjustment for the next grasp. Re-grasping thus is performed iteratively until our model identifies a stable grasp. Finally, we demonstrate extensive experimental results on grasping a large set of novel objects using tactile sensing alone. Furthermore, when applied on top of a vision-based policy, our re-grasping model significantly boosts the overall accuracy by $10.6\%$. We believe this is the first attempt at learning to grasp with only tactile sensing and without any prior object knowledge. For supplementary video and dataset see: \href{http://www.cs.cmu.edu/afs/cs/user/amurali/www/projects/GraspingWithoutSeeing/}{cs.cmu.edu/GraspingWithoutSeeing}.

\end{abstract}

\section{Introduction}
Consider the task of grasping a slippery glass bottle. We use vision to determine the object's location and its properties such as shape. Based on these estimates, we can even plan how to approach and make contact with the bottle. However, not until we get tactile feedback by touching, can we adjust our hands for a reliable grasp. In many cases, the hand completely occludes the object after contact, severely diminishing the use of hand-eye coordination; yet in all these cases we humans are invariably successful in grasping the objects. In fact, we are even capable of grasping objects solely based on touching. A good example is when we probe around on a nightstand for our phone. Haptics and the sense of touch plays a vital role in grasping. Yet, most of our currently existing grasping algorithms  primarily builds on visual sensing (RGB-Depth or laser scanners). In fact, in the recent Amazon Picking Challenge, only one of 26 teams used a tactile sensor~\cite{apc2017}. Can a robot learn to grasp solely based on touching and without even using vision? More importantly, can the robot incorporate both visual inputs and tactile feedback for robust grasping?

\begin{figure}
  \begin{center}
    \includegraphics[width = 0.9\linewidth]{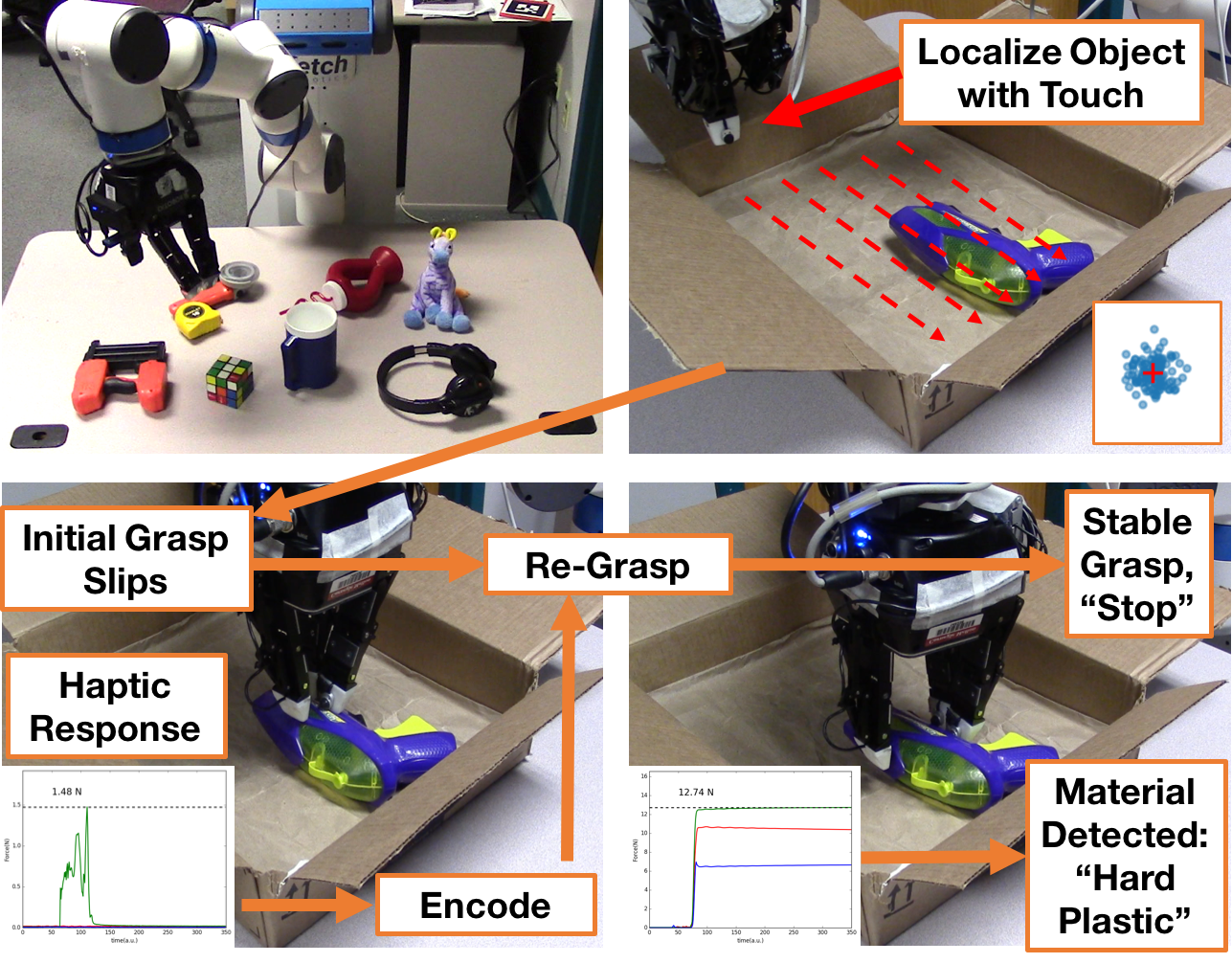}
  \end{center}
  \vspace{-1.2em}
  \caption{Our Fetch robot learns to localize and grasp a novel object of unknown shape from just tactile sensing. Our method estimates the target's location by touch-probing the workspace (top right), and establish an initial grasp (bottom left). We then learn to extract features from haptic feedback, and predict how to adjust the grasp (bottom right). This re-grasping process is repeated until our method identifies a stable grasp. }
  \vspace{-2.0em}
\label{fig:cover}
\end{figure}

Sensory inputs affect the success of a grasp in all stages: \textit{localization} of the object, \textit{planning}\footnote{Grasp planning refers to both analytic and data-driven techniques.} of the grasp control parameters (gripper pose, approach direction, etc.) and the \textit{execution} of the grasp on the robot. Vision-based methods, such as object detection, segmentation and point cloud registration, are widely used for localization. Without using visual sensing, tactile exploration has demonstrated promising results on locating objects and estimating their 6 DOF poses~\cite{javdani2013efficient,saund2017touch,petrovskaya2011global,koval2013pose,pezzementi2011object,kaboli2017tactile}. However, haptics has rarely been considered in the context of grasping beyond simple, individual objects. Recently, there has also been tremendous progress in data-driven grasp planning methods, namely in learning grasp policies from RGB-D images~\cite{watter2015embed,pinto2016supersizing,levine2016learning, mahler2017dex}. But most of these approaches ignore haptic feedback during execution. In fact, tactile sensing has been previously used for grasp execution, for instance in assessing grasp stability~\cite{bekiroglu2011assessing,dang2014stable, calandra2017}, and thus enabling the hand to adjust its posture and position online~\cite{romano2011human,dang2011blind,hsiao2010contact,dang2013grasp}. Nonetheless, these methods assume either the initial grasp or the object information is inferred with vision, with few exceptions~\cite{felip2012}. Felip et al.\ \cite{felip2012} presented a full system for tactile grasping using hand-crafted rules. In such light, no general learning framework exists for a complete grasp (localization, planning and execution) using solely touch sensors.

In this work, we present the first general framework for learning to grasp with only tactile sensing and without prior object knowledge. Our goal is to scale to a diverse set of unknown objects. To this end, we focus on 2D planar grasps of a single object. To start with,  we design a {\it localization module} to obtain an approximate location of the object. Intuitively, we control the robot to sequentially ``touch-scan'' the grasp plane until hitting the object and we use a particle filter to aggregate the measurements and track the target location. 

With all the uncertainty of object location, tactile sensing and kinematics, how can the robot reliably grasp the object? Our core idea is to treat grasping as a multi-step process with error recovery. Specifically, we propose a {\it re-grasping module} that refines the initial grasp with multiple re-trials. To extract rich meaningful features for the re-grasping task, we use a recurrent auto-encoder to learn an unsupervised representation from all the unlabelled data. These features are then fed to another neural network that simultaneously estimates grasp stability, and predicts the adjustment for the next grasp. Our framework will iterate on the grasps until our network estimates a high chance of success or the number of trials reaches a predefined limit. 


Our high-capacity deep network requires a large-scale tactile dataset for training, which is missing in the community. We have thus created a new dataset of grasping with both tactile and visual sensing. Specifically, we record images, haptic measurements as the robot gripper encloses its fingers on an object, high-level re-grasp actions sent to the motion planner and labels of whether an object has been successfully grasped. Our publicly available dataset includes $7.8$K interactions with $52$ unique objects with material labels. We hope that it will serve as a major resource for future research on visio-haptic manipulation.

Our method is trained using our dataset, and tested on $20$ unseen objects. We systematically vary components of our framework and benchmark the performance. First, we show that our unsupervised representation learning produces rich tactile features for a variety of passive (material recognition) and active (re-grasping) tasks. Next, we show that haptic based re-grasping improves a baseline policy, with the ground truth object location provided by vision-based localization. Finally, with touch based localization, our full method achieves a grasping accuracy of $40.0\%$ using tactile sensing alone. We believe this is one of the first results of grasping a large set of unknown objects without seeing. Furthermore, we explore combining haptic and visual sensing for robust grasping. Our results indicate that our multi-step re-grasping with tactile feedback 1) improves the robustness of grasp execution and 2) offers an easy plug-in for existing grasp planning methods.



\section{Related Work}
Grasping is one of the fundamental problems in robotic manipulation and we refer readers to recent surveys~\cite{bicchi2000robotic, tegin2005tactile, bohg2014data}.

{\bf Vision Based Grasping.} Visual perception has been the primary modality for sensing, grasp planning and execution. Several work on model-based grasping make use of visual information like point clouds/images to estimate physical properties of objects (e.g., shape~\cite{miller2003automatic} or pose~\cite{collet2009object}), and finally to generate control commands for grasping. Sensing detailed physical properties from visual inputs can be exceedingly challenging, and might not be necessary for finding desired controls. Therefore, recent papers have focused on learning-based approaches~\cite{saxena2006learning,watter2015embed}. These methods directly map input visual data to the control signals for open-loop grasping. Recently, a lot of progress has been made in this direction by using deep models~\cite{levine2016end,pinto2016supersizing,levine2016learning, mahler2017dex}. However, using visual inputs alone leads to errors such as slippage due to low-friction or wrong grasp location due to self-occlusion.


{\bf Tactile Exploration.} In contrast, humans make great use of tactile signals for grasping and can even grasp unknown objects without using visual sensing~\cite{johansson2009coding}. Therefore, recent work in robotics has also explored the use of haptics for sensing an object's shape, pose, location or attributes~\cite{chu2013using,spiers2016,yuan2017shape}. For example, if the location of an object is known, the shape can be estimated by actively touching its parts~\cite{yi2016active,meier2011probabilistic}. Similarly, given the 3D models of objects, several recent work seek to infer the 6DOF pose of the objects with a series of information-gathering actions~\cite{petrovskaya2011global,saund2017touch,javdani2013efficient}. However, these results have neither been considered for the task of grasping nor can generalize to unknown objects. The most relevant work are from~\cite{pezzementi2011object} and~\cite{kaboli2017tactile}. Pezzementi et al.\ \cite{pezzementi2011object} built occupancy grid mapping using tactile sensing of unknown 2D objects. Kaboli et al.\ \cite{kaboli2017tactile} proposed a pre-touch strategy to localize novel objects in a 3D workspace. These work are similar to our touch localization step, yet they failed to complete the full pipeline of touch-based grasping. 

\begin{figure*}[t]
\centering
\includegraphics[width = 0.92\linewidth]{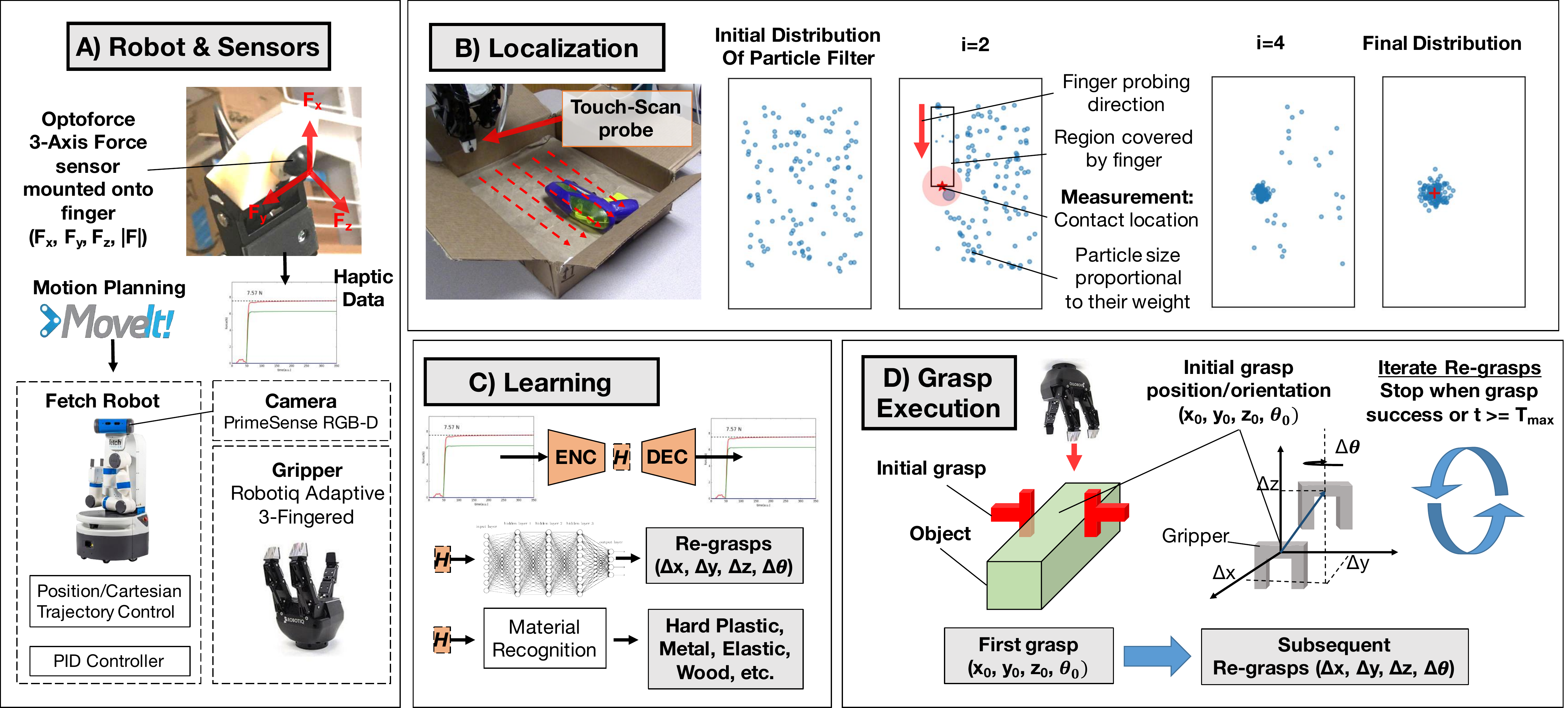}
  \vspace{-0.1in}
\caption{Overview of our system and approach. (a) Our robot and sensors: We equip a fetch robot with a Robotiq gripper and additional sensor packages. Our sensors include force sensor on the fingers of the gripper and RGB-D cameras on the head of the robot; (b) Our touch based object localization: We touch-probe a 2D grasp plane of the workspace, and use particle filtering to aggregate evidences of the object's location. An initial grasp is established given an estimate of the object's 2D location. (c) Our unsupervised learning scheme for haptic features: We learn to represent haptic data during grasping using an conditional auto-encoder. The learned features are fed into our re-grasping model to correct the initial grasp. (d) Our re-grasping model: Based on haptic features from current grasp, we estimate grasp stability and predict how to adjust the grasp. A new grasp is generated by applying the adjustment to the current grasp. This process repeats until our method predicts a stable grasp.}
  \vspace{-1.5em}
\label{fg:giant_figure}
\end{figure*}

{\bf Re-grasping with Tactile sensing.} Haptic feedback is widely used for closed-loop control when executing a grasp, also known as re-grasping. Early work~\cite{fearing1986simplified} focused on analytical solutions for 2D planar grasp given ideal tactile sensing of a known object shape. For real world tactile data, hand crafted rules can be highly effective if object shape is known~\cite{hsiao2010contact}. Several recent works addressed the task of re-grasping or assessing grasp stability without prior object knowledge~\cite{romano2011human,bekiroglu2011assessing,chebotar2016self,dang2013grasp,dragiev2013uncertainty, calandra2017}. However, they all rely on a good initial grasp given by another sensor modality. The most relevant work are from~\cite{dang2011blind,koval2016pre} and~\cite{felip2012}. Based on tactile feedback, Dang et al.\ \cite{dang2011blind} learned to predict grasp stability~\cite{dang2014stable}, which is further used to guide grasping. Their method can generalize to unknown objects but requires accurate object locations. Moreover, their approach only used simulated data with hand-designed features. Koval et al.\ \cite{koval2016pre} utilized haptic feedback to learn both pre and post contact push-grasping policies. Their method accounts for inaccurate sensing of object location and pose, yet is limited to objects with known shapes. Conversely, our method learns tactile based re-grasping policies with neither prior knowledge of the object (shape/physics) nor necessarily a good initial grasp. In addition, our approach makes use of large-scale real-world visual and haptic data to learn how to grasp. Moreover, Felip et al.\ \cite{felip2012} presented a full tactile grasping pipeline (exploration and re-grasping) with a wrist force-torque sensor, fingertip tactile sensors and a fully actuated multi-fingered gripper. They used a set of hand-crafted rules/features and demonstrated success on a small set of novel objects. Conversely, our tactile perception modules are learned from data and only uses the fingertip tactile sensors. We show that our learned model can be applied to successfully grasp a larger set of novel objects, including deformable and elastic ones. 

{\bf Grasping Datasets.} Alongside algorithmic developments, large-scale datasets have fueled the success of learning to grasp~\cite{levine2016end,pinto2016supersizing}. However, when it comes to haptic datasets, there have been only few attempts such as~\cite{chebotar2016bigs,erickson2017semi}. These datasets either focus on passive tasks e.g., material recognition~\cite{erickson2017semi}, or are limited to grasping a small set of $2$-$3$ objects with a small number of trials~\cite{chebotar2016bigs, chebotar2016iser}. As part of our effort, we created the first large-scale grasping dataset with both tactile and visual sensing to facilitate future research of visio-haptic grasping. As a result, our work is also deeply intertwined with the unsupervised learning of tactile feature representations. Previous work has primarily used hand-crafted features for haptic data~\cite{hsiao2010contact}. Schneider et al.\ \cite{schneider2009} constructed haptic features using bag-of-words. Madry et al.\ \cite{madry2014st} explored unsupervised learning of haptic features using sparse coding.  The learned representation has been shown effective for re-grasping~\cite{chebotar2016self}, though it is intended for a specific class of sensors providing a matrix/image of tactile responses. We propose a novel method for learning haptic features using a deep recurrent network similar to~\cite{sung2017learning}.

\section{Dataset}
In this section, we present the effort on creating our visio-haptic dataset for grasping. Large-scale haptic dataset for grasping is important for learning high capacity deep models. Unfortunately, this kind of dataset is missing in the community. We seek to bridge this gap by collecting a new grasping dataset that includes both visual and haptic sensor data. Specifically, our dataset consists of $7800$ grasp interactions with $52$ different objects. Each grasp interaction lasts for $3.5$-$4$ seconds and is recorded with:

\begin{itemize}
    \item \textbf{RGB Frames:} We capture images of four specific events of a grasping: for the initial scene, before, during and after grasp execution. These images have a resolution of 1280x960. 
    \item \textbf{Haptic Measurements:} Tactile signals are measured by force sensors mounted on each of the three fingers of the gripper. The sensor measures the magnitude ($F$) and the direction of forces ($F_x, F_y, F_z$) at $100$Hz.
    \item \textbf{Grasping Actions and Labels:} We record the pose of all 2D planar grasps, including the initial grasp $(x_0, y_0, z_0, \theta_0)$ and subsequent re-grasps $(x_t, y_t, z_t, \theta_t)$. We also record whether the re-grasp succeeded.
    \item \textbf{Material Labels of Objects:} We label material categories (7) for each object, including metal, hard plastic, elastic plastic, stuffed fabric, wood, glass and ceramic.
\end{itemize}
\noindent 

\textbf{Data Collection.} To collect this dataset, we sample and execute a large set of grasps. The robot will lift up objects and automatically detect successful grasps. A major issue with this data collection process is how we can get more successful grasps. It is easy to collect failure cases by applying random grasps but it is difficult to collect successful grasps, which are critical for learning. To address this issue, we used an existing vision based grasping policy to sample an initial grasp from a pre-learned visual grasping policy~\cite{murali2017cassl}. We collect two sets of data and combine them to form our final dataset. The first set includes all $52$ objects with $50$-$55$ initial grasps. Each initial grasp is followed by a single random re-grasp. The grasps in this set have a higher rate of success. On the other hand, our second set contains a subset of $7$ objects covering different types of materials. For each object in this set, we sample $80$-$100$ initial grasps, and allow $2$-$3$ random re-grasps, resulting in a higher failure rate. 

\textbf{Dataset Statistics.} Overall, our dataset includes more than $30$K RGB frames and over $2.8$ million of tactile samples from $7800$ grasp interactions of $52$ objects. We provide grasping actions and labels for each interaction, as well as material labels for each object. To the best of our knowledge, this is by far the largest dataset for vision-haptic grasping. Our dataset is publicly available at: \href{http://www.cs.cmu.edu/afs/cs/user/amurali/www/projects/GraspingWithoutSeeing/}{cs.cmu.edu/GraspingWithoutSeeing}. 


\section{Overview}
We present an overview of our framework in Fig~\ref{fg:giant_figure}. Our goal is to reliably grasp a target object using just fingertip tactile sensors and without knowing the location, pose or shape of the object. Similar to previous works, our framework has two main stages: grasp planning and grasp execution. For planning, we make use of particle filtering to localize an object based on a sequence of touch-probing. For grasp execution, we learn to iteratively adjust the grasp based on haptic feedback, using a deep neural network. Unlike other work in robot learning \cite{levine2016end} which learn torque control, we infer position control commands and use a motion planner to reach that configuration. We also explore the benefit of applying our re-grasping model on top of a vision based grasping policy. Our methods for planning and execution are detailed in Section~\ref{sec:planning} and~\ref{sec:regrasping}, respectively.

{\bf Platform.} We implement our method on a real world robotic platform---a research edition of Fetch mobile manipulator~\cite{fetch2016}, equipped with a 3-Finger adaptive gripper (Robotiq). We use ROS~\cite{quigley2009ros} and position control with the Expanding Space Tree (ESTk) motion planner from MoveIt to generate collision-free trajectories for the robot. For haptic sensing, we mount a 3-Axis Optoforce sensor onto each of the three Robotiq fingers. We made sure this mounting is rigid by using customized 3D-printed fixtures (see left panel of Fig~\ref{fg:giant_figure}). For vision, we use a PrimeSense Carmine 1.09 short-range RGB-D camera mounted on the robot's head. Note that visual data is not used in our method, except when we explore combining RGB frames from PrimeSense with haptic sensing for grasping.



\section{Initial Grasp from Touching}\label{sec:planning}
We present our method for grasp planning. Traditionally, the goal of planning is to generate a good initial grasp of a target object. This usually requires the robot to sense the physical properties of the object, such as shape or pose. This is especially challenging with tactile sensing alone. Nevertheless, our key observations are that 1) we can infer a rough location of the object by probing the grasp plane and hitting the target multiple times; 2) even a poor initial grasp is often sufficient for successful grasping, if we allow the robot to correct the grasp a few times using haptic feedback. Thus, we propose a simple method for grasping. We first localize the object by touching, and then generate a random initial grasp. We will show that this method can be highly effective when combined with our learning based re-grasping policy. 

\subsection{Particle Filter for Touch Localization}
The core of our grasp planning is a simple touch-based localization method using contact sensing. We consider the task of grasping a single target object within a known workspace--in our setting a constrained packaging box in which the object could be in any pose. In this case, we control the robot to line-scan a fixed $2$D plane of the workspace using one of its fingers, which functions as a touch probe. The probe moves in a cartesian path until it detects a contact (defined by a threshold on the magnitude of force). Our method makes multiple contacts and uses particle filtering to infer the object's location $x\in R^2$ on the $2$D plane. 

The choice of a particle filter is tailored for our problem, as our contact measurement is highly non-linear and lacks analytic derivatives. Particle filters are a non-parametric formulation of the recursive Bayes filter:
\begin{equation}
\small
bel(x_{t}) = \eta p(z_{t}|x_{t}, u_{t}) \int p(x_{t}|x_{t-1}, u_{t}) bel(x_{t-1}) dx_{t-1}
\label{eq:bayesfilter}
\end{equation}
The belief $bel(x_{t})$ is approximated using a finite set of particles $X_{t}=\{x_{t}^{[i]}\}_{i=1}^n \app bel(x_{t})$. $x_t$ above denotes the target location at time $t$, $u_t$ is the line scan action and $z_t$ the contact sensing measurement. The touch-localization framework is summarized in Algorithm~\ref{alg:touchlocalization} and the detailed mechanisms of the particle filter could be found in \cite{particlefilter2001}. At the end of touch-scanning, the centroid of the resampled $X_{N_{SCANS}}$ particles is returned as the final estimate of the target object's location. 

\begin{algorithm}[H]
\small
    $X_{0}$ $\leftarrow$ Uniform random samples \; \\    
    \For{t = 1:$N_{scans}$}
    {
        $X_{t}$ $\leftarrow$ $\O$ \; \\
        Run linear scan $u_{t}$ to get observation $z_{t}$ \; \\
        \For{i = 1:$N_{particles}$ in $X_{t-1}$}
        {
            Sample from motion model $x_{t}^{[i]} \app$ $p(x_{t}|x_{t-1}^{[i]})$ \; \\
            Update measurement $w_{t}^{[i]}$ $\leftarrow$ $p(z_{t} | x_{t}^{[i]}$, $u_{t})$ \; \\
            $X_{t}$ $\leftarrow$ $\{ x_{t}^{[i]} \}$ $\cup$ $X_{t}$\; \\
        }
        $X_{t}$  $\leftarrow$ Resample($X_{t}$, $w_{t}$) \; \\
    }
    \textbf{return:} mean($X_t=\{x_t^{[i]}\}$) $\rightarrow$ object location \;
    \caption{Touch localization using Contact Sensing}\label{alg:touchlocalization}
\end{algorithm}
\vspace{-1.2em}

We present details of our measurement and motion models. 
\begin{itemize}
\item \textbf{Motion model}: Touching the object might change its location. This displacement is usually small, yet is determined by how the robot moves ($u_t$), and the physical properties of the object and its environment. We simplify the motion model by assuming a Gaussian distribution independent of $u_t$: $p(x_{t} | x_{t-1}, u_{t}) = \mathcal{N}(x_{t-1}, \sigma^2 I)$, where $\sigma$ is a small noise.
\item \textbf{Measurement model}: Our measurement model tracks physical occupancy of probed locations. Any location on the $2$D plane can be either free space (no contact) or occupied by the object (contact). We either increase (occupied) or decrease (free space) the weights of particles that lies within the vicinity (a sphere of radius 2.5cm for our experiments) of the location. An example is shown in Fig~\ref{fg:giant_figure}, where particles in swept area of the probe are down-weighted and particles near the contact point (red circle) are up-weighted.
\end{itemize}

Once we estimate the target location, our next step is to generate a grasp. Without prior object information, we select a grasp by randomly sampling from the rest of the parameter space. Executing such a grasp is highly likely to fail, as this sample can be far away from feasible grasps. Somewhat surprisingly, we will show that this random policy can produce a successful grasp, if we allow the robot to re-grasp a few times and adjust its controls each time based on tactile feedback.

\section{Grasp Execution via Re-grasping}\label{sec:regrasping}

\begin{figure}[t]
  \begin{center}
    \includegraphics[width = 0.9\linewidth]{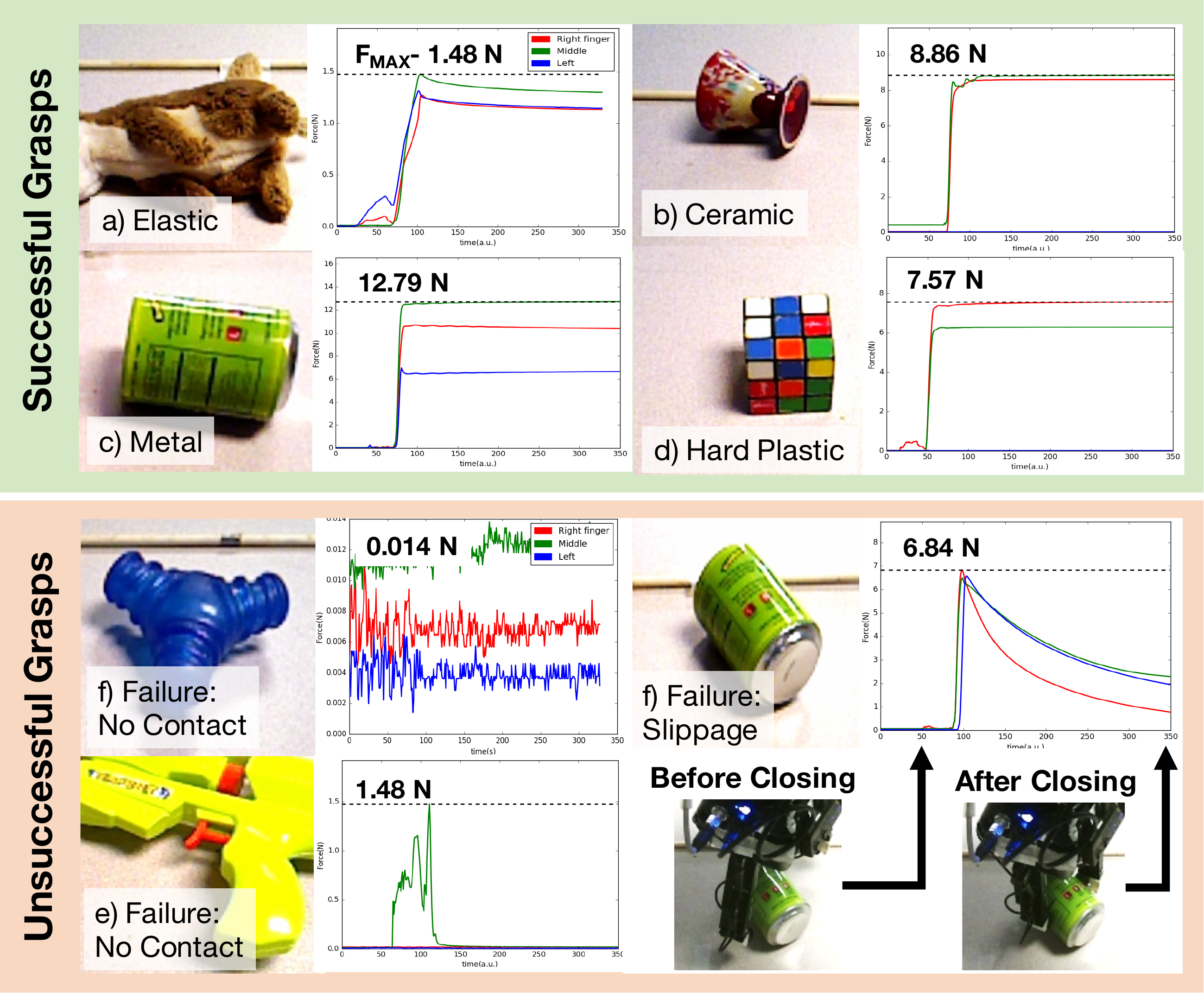}
  \end{center}
  \vspace{-1.2em}
  \caption{Tactile response from both successful and failed grasps. These grasps are from objects with varying shape/material/compliance properties. We plot the time series of force magnitude from our sensors  on three fingers (red: right, green: middle, blue: left). The maximum force during grasping is also displayed. We record signals before and after the gripper closes (shown in bottom). These signals contain important information about the object (e.g., material, shape) and the grasping (e.g., grasp stability). And we explore using them to estimate how to correct a previous grasp.} 
  \vspace{-1.5em}
\label{fig:haptics}
\end{figure}

Given a noisy object location and a randomly selected grasp, how can the robot reliably grasp the object? To address this question, let us first look at what is measured by haptic sensors during grasping. Fig~\ref{fig:haptics} shows haptic responses during the task of grasping. It is evident that these signals encode important information about the object in contact. For example, the magnitude of force implies the material of the object. And the temporal force variation across three fingers indicates the shape. These signals also capture critical aspects about the grasping. For example, we can predict the stability of the grasping by tracking the temporal structure of signals before and after contact. Therefore, we hypothesize that these tactile signals can be used to correct the initial grasp.

We will demonstrate that this is indeed possible if we consider grasping as a multi-stage process, and allow the robot to re-grasp a few times. Each new grasp is generated by adjusting a previous one using haptic feedback. Re-grasping thus helps to reduce the uncertainty of sensing. To this end, we propose a learning based approach for tactile based re-grasping. Our method learns representations from haptic data, estimate the grasp stability and predict the adjustment for next grasp, all using deep models. We now present our methods on haptic feature learning and tactile based re-grasping. 

\subsection{Learning Haptic Features}
\label{sc:haptic_features}
\begin{figure}[t]
\centering
\includegraphics[width =0.85\linewidth]{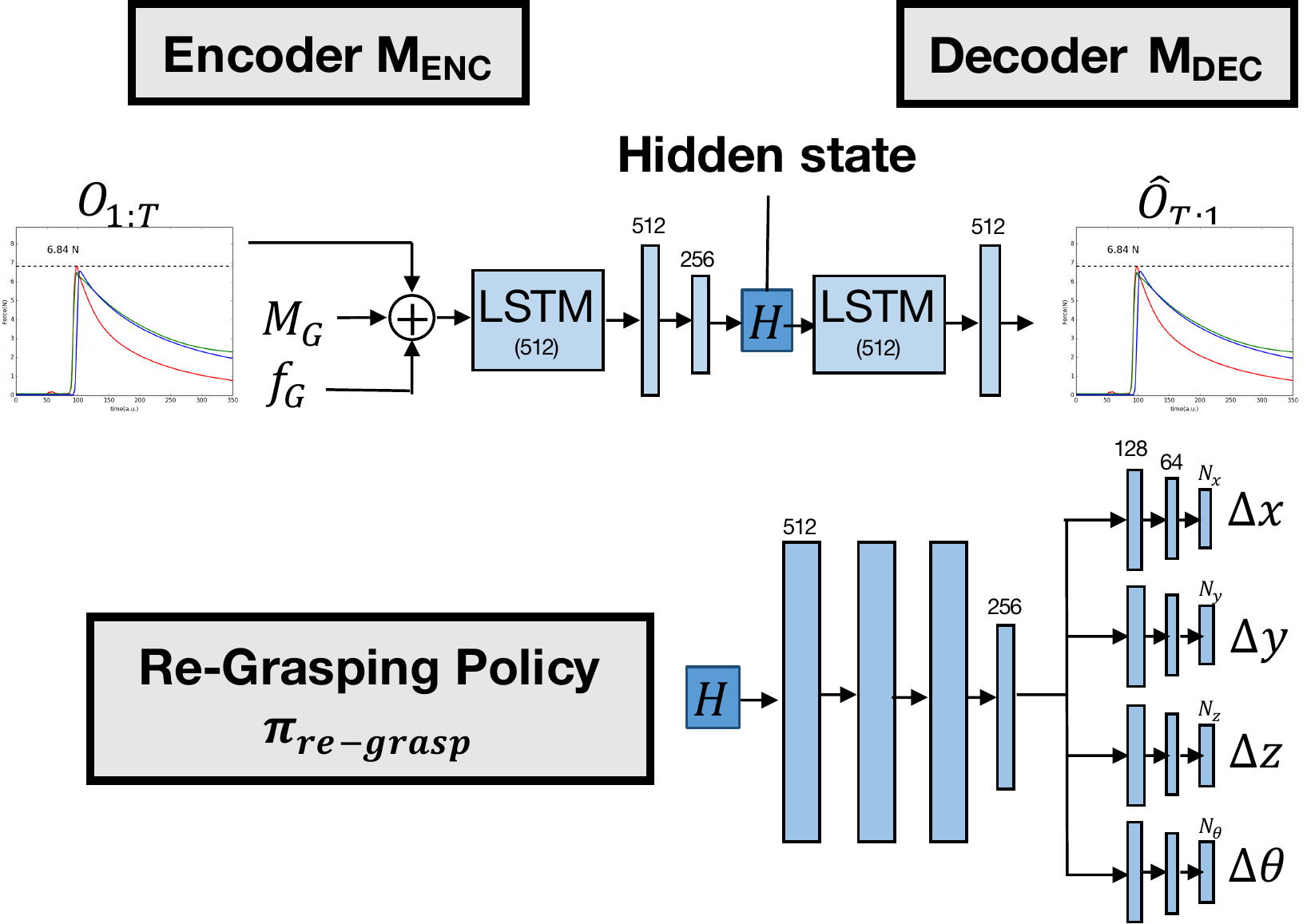}
\vspace{-1.0em}
\caption{Network architectures for learning haptic features (top) and re-grasping policy (bottom). Our conditional auto-encoder $M_{ENC}$--$M_{DEC}$ learns to reconstruct haptic data using both haptic signals and applied gripper control. We treat the learned latent space $H$ as features for learning re-grasping policy $\pi_{re-grasp}$. Our re-grasping policy maps the hidden representation $H$ to the adjustments of planar grasping parameters ($\Delta x, \Delta y, \Delta z, \Delta \theta$) (4D). These high level parameters are then executed using the motion planner to generate a new grasp.}
\vspace{-1.5em}
\label{fg:ConditionalEncoder}
\end{figure}

The next question is how do we learn a generalized representation of haptic data? Should we use hand-designed features or some task-specific representation? Raw tactile signals are in the form of a time series, with a low dimensional vector at each time step. Since they do not encode much global information compared to modalities like vision, it is challenging to consider haptic data without the context of the robot control applied. Therefore, what we need to learn is a conditional representation and to this end, we trained a conditional auto-encoder model over the haptic signals, shown in Fig~\ref{fg:ConditionalEncoder}. Both encoder and decoder in our model have a recurrent architecture (LSTMs~\cite{hochreiter1997long}). Our encoder $M_{ENC}$ takes a sequence of haptic data and control signals as inputs, and encodes them into a low dimensional latent space $H$. Our decoder $M_{DEC}$ reconstructs the input haptic data from the latent space $H$.

By conditioning the reconstruction on control actions, the network must learn to embody the temporal structure of haptic data within the motion of the robot during grasping. This will allow us to re-use $H$ to present haptic and control signals for re-grasping. Note that the learning is unsupervised in nature and does not require manual labeling. 

More specifically, our haptic signals, denoted by $O = \{O_{t}\}$, include a $12$D vector for each time step from all three fingers. Our control signals include the configuration of the gripper: $f = \{f_{t}\}$ and $m = \{m_{t}\}$. $m_{t}$ is the mode of the adaptive gripper. $m$ describes the angle between the fingers, and has categorical values of ``pinch'', ``normal'' and ``wide angle''. The under-actuated gripper fingers have three links each but only one DOF as $f_{t}$. $f_{t}$ is valid when the gripper has been fully enclosed on the object. If no object was enclosed (grasp failure), $f_{t}$ will take the maximum possible value. We use $L$2 loss and stochastic gradient descent for training. For feature extraction, we discard the decoder $M_{DEC}$ and only use the encoder $M_{ENC}$ to extract the hidden state $H$ from a fixed size time window ($3$ seconds).

\subsection{Learning to Re-grasp}
We consider a multi-stage grasping problem, where each grasp is conditioned on the previous one. Formally, given a current grasp $g$, we measure the haptic data $O$ and grasp configuration parameters ($m$, $f$) and encode them into $H=M_{ENC}(O, m, f)$. $H$ is the hidden state that captures the haptic responses of the current grasp. Next, we learn the corrective action $\Delta g = \pi_{re-grasp} (H)$ that leads to better grasp stability and the architecture is shown in Fig \ref{fg:ConditionalEncoder}. At the same time, we learn a score function $p = M_{stability}(H)$ to predict the grasp stability, which determines the empirical probability of grasp success. The score function $M_{stability}(H)$ is a simple feedforward networks with 5 fully connected layers of size $(512, 512, 256, 128, 64)$ and a final sigmoid function to estimate the probability. When testing, we iteratively apply the predicted $\Delta g$ to current grasp $g$. We execute $g+\Delta g$ until $M_{stability}$ predicts a high rate of success. Algorithm~\ref{alg:gwos} summarizes our method. 

\begin{algorithm}[H]
\small
    Localize object with vision/touch \;
    
    Sample $g_{1}$ from $\pi_{vision}$/$\pi_{random}$ \;
    
    Execute $g_{1}$ on robot \;
    
    Collect first haptic measurement $O_{1}$ \;
    
    \For{i = 2:$T_{max}$}
    {  
        Encode $H_{i-1}$ $\leftarrow$ $M_{ENC}(O_{i-1})$ \;
        
        Compute $p_{i-1}$ = $M_{stability}$($H_{i-1}$) \;
        
        \eIf{$p_{i-1}$ $>$ $p_{threshold}$}
        {
            break \:
        }
        {
            Compute $g_{i}$ = $\pi_{regrasps}$($H_{i-1}$) \;
            
            Execute $g_{i}$ on robot \;

            Collect haptic measurement $O_{i}$ \;
        }
    }
\caption{Grasping Without Seeing}\label{alg:gwos}
\end{algorithm}
\vspace{-1.2em}

Our output action $\Delta g$ is parameterized by the change of the gripper's position (-0.025m $\leq$ ($\Delta x$, $\Delta y$, $\Delta z$) $\leq$ 0.025m) and orientation ($-\pi/4 \leq$ $\Delta \theta$ $\leq \pi/4$). $\Delta g$ is thus a $4$D vector. Given that the haptic measurement is only relevant in the local neighborhood of the current grasp, we constrain the range of these parameters to small adjustments tailored to our setting. During data collection, continuous values of the re-grasp ($\Delta x$, $\Delta y$, $\Delta z$, $\Delta \theta$) are sampled randomly. However, for the deep network we use a dicretized output space. Specifically, we discretize each control dimension into $5$ bins. Thus, the learning of the policy function $\pi_{re-grasp} (H)$ is similar to multi-way binary classification. 
\begin{equation}
\small
L = \sum_{i=1}^{K} \sum_{k=1}^{B} \sum_{j=1}^{D(i)} \delta(k, u_{i,j}) \cdot \text{Cross-Entropy}(\sigma(y_{ij}^{final}),\hat{y}).
\label{eq:loss}
\end{equation}
Eq~\ref{eq:loss} shows our loss for learning our policy function. $\hat{y}$ corresponds to the success/failure label while $y_{ij}^{final}$ is the final dense layer before the sigmoid. $D(i)=5$ gives the number of discretized bins for control parameter i, K (=4) is the number of control parameters, B is the batch size and $\sigma$ is the sigmoid activation. $\delta(k, u_{i,j})$ is an indicator function and is equal to 1 when the control parameter i $u_{i,j}$ corresponding to bin j is applied. The learning rates for $\pi_{re-grasp}$, $M_{ENC}/M_{DEC}$, $M_{stability}$ are 5e-7, 1e-5 and 5e-5 respectively. All models are trained with ADAM optimizer~\cite{kingma2014adam} for around 20 epochs. The networks and optimization are implemented in TensorFlow~\cite{abadi2016tensorflow} and Keras. Similarly, $M_{stability}$ is learned using a cross-entropy loss.


\subsection{Improving Vision-Based Grasping with Re-grasping}
Finally, for our experiments  we also explore incorporating the haptic re-grasping module with vision based grasping. In practice, any vision-based policies could be used~\cite{pinto2016supersizing, mahler2017dex, levine2016learning}. We adapt a variant of~\cite{murali2017cassl} (hereafter denoted as $\pi_{vision}$). $\pi_{vision}$ is used to generate an initial grasp, followed by our re-grasping model. We also use this policy to collect our dataset. We sample control parameters from $\pi_{vision}$ that are more likely to produce a success grasp to increase the number of successful grasps in our dataset. 

Specifically, five control parameters are inferred from the object's image $I_{obj}$ : $x_{pixel}$, $y_{pixel}$, $\theta$, $M_{G}$, $h_{G} \app \pi_{vision}(I_{obj})$. $x_{pixel}$ and $y_{pixel}$ are the 2-D grasp locations in image plane (converted to 3-D coordinates $x_{G}$ and $y_{G}$ with a calibrated depth camera). $\theta$ is the angle of the gripper about the vertical axis in a planar grasp (similar to~\cite{pinto2016supersizing}). $M_G$ is the configuration of the gripper, which is also used for our learning of haptic features. And $h_G$ is estimated height of the object from depth sensing. For both testing and data collection, we sampled $N_{patches}=40$ parameters from $\pi_{vision}$ and chose the command $u_{i}$ for each control dimension $i$ by $u_{i} = \argmax_{j} \pi_{vision}(I_{obj}, u_{ij})$.

\section{Experimental Evaluation}

We now present our experimental results. Our experiments are divided into two parts. First, we evaluate the learned haptic features for two key tactile perception tasks of material recognition and grasp stability estimation. We compare against state-of-the-art haptic feature extraction methods, and benchmark the choice of classifiers. Second, we test our tactile based grasping framework. We report results for our re-grasping module, tactile-only grasping, and visio-haptic grasping. 

\begin{figure}[t]
\centering
\includegraphics[width = \linewidth]{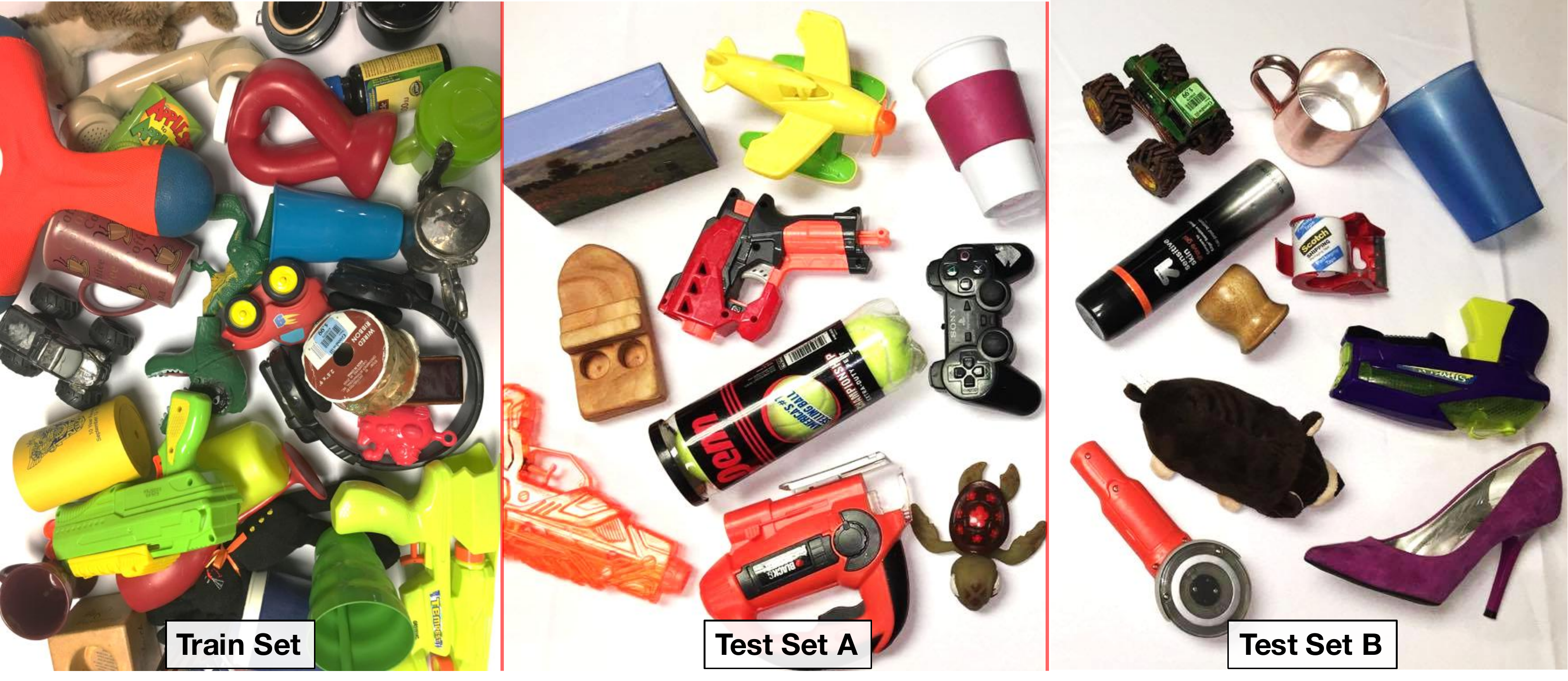}
\vspace{-1.2em}
\caption{Our test set of objects. These objects were not in the training data. We divide our test set into two parts. Set A contains slightly harder objects to grasp (such as the red and orange toy guns) compared to Set B.
}
\label{fg:test_set}
\vspace{-1.5em}
\end{figure}

\textbf{Test Set for Grasping.} To evaluate our grasping framework, we physically test grasping methods on a set of novel objects. We measure the grasp accuracy averaged over multiple trials per object as our evaluation criteria. This test setting is very challenging: testing objects are not presented in the training set and thus have not been seen by neither our $\pi_{re-grasp}$ model nor $\pi_{vision}$. Our testing set is divided into two parts, as shown in Fig~\ref{fg:test_set}. Each set consists of $10$ different objects. Set A is more difficult than Set B, as it contains objects with more complex geometry, heterogeneous material distribution (e.g., plastic toy guns and stapler) and articulations. This test set is also used for grasp stability estimation.

\subsection{Learning Haptic Features}

Our first experiment tests our haptic feature learning scheme. Our decoder achieves a reconstruction error (L2 norm) of $0.81$ and $1.1$ on the training set and our held-out testing set ($10\%$ of the recorded data), respectively. This error (around 1 Newton of force) is reasonable when compared to $\app 0.2$ Newton sensing noise from our force sensor. To further evaluate the learned haptic features, we consider two key tasks in tactile perceptions: (1) material recognition; and  (2) grasp stability estimation. And we consider different combinations of haptic features and classifiers for both tasks.

\textbf{Tactile Features.} We compare our learned haptic features with two other baselines representation learning methods.
\begin{itemize}
    \item \textbf{Auto-encoder.} This is our haptic features learned using a unsupervised recurrent auto-encoder. Once learned, only the encoder is used to extract features.
    \item \textbf{Sparse Coding.} This is a variant~\cite{roberge2016sparsecoding} of ST-HMP features~\cite{madry2014st}. These features are learned using dictionary learning and sparse coding on the spectrogram of 1D time series of tactile signals. Note that directly using ST-HMP is not feasible for us, as it requires 2D tactile images.
    \item \textbf{Hand Crafted.} This is from~\cite{spiers2016}, where raw signals from three specific events (before contact, when the finger closing movement is stalled due to object-finger contact, after the fingers are in equilibrium) are extracted. 
\end{itemize}

\textbf{Choice of Classifiers.} We further vary the classifiers used for both material recognition and grasp stability estimation. 
\begin{itemize}
    \item \textbf{Deep Network.} We train a five-layer neural network with cross entropy loss for classification. 
    \item \textbf{SVM.} This is a linear classifier trained with hinge loss. 
\end{itemize}

\subsubsection{Material Recognition}
The task is to classify $7$ different materials in our dataset using tactile signals during grasping. All features are learned from the full training set, as no supervision is required. Our classifiers are trained on a subset of the training set (80\%) and tested on the held out testing set (the remaining 20\%). We report average class accuracy. The results are presented in Table~\ref{tb:MaterialClassification}.

\begin{table}[H]
\footnotesize
\centering
\caption{Results of Material Recognition}
    \vspace{-1em}
\label{tb:MaterialClassification}
\begin{tabular}{c|cc}
\hline
\multirow{2}{*}{Feature Type} & \multicolumn{2}{c}{Accuracy  (\%)} \\
   & Deep Network      & SVM    \\
\hline
Auto-encoder (Ours)                & \textbf{42.86}       & 40.68       \\
Sparse Coding \cite{roberge2016sparsecoding, madry2014st}                    & 36.35       & 35.93     \\
Hand Crafted \cite{spiers2016}                    & 33.50       & 33.66     \\
\hline
\end{tabular}
\vspace{-1.5em}
\end{table}

\begin{figure}[t]
  \begin{center}
    \includegraphics[width = 0.75\linewidth]{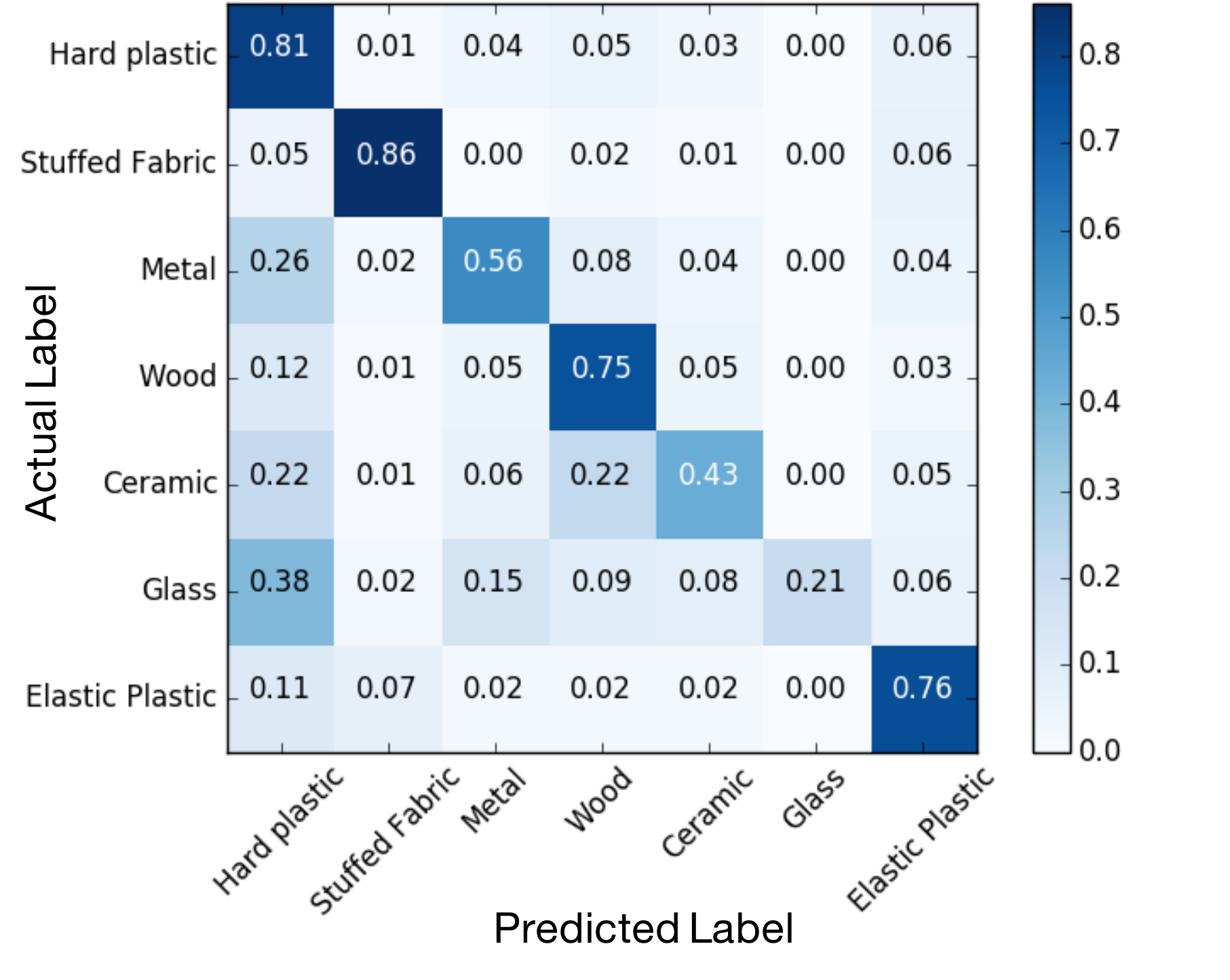}
  \end{center}
  \vspace{-1.5em}
  \caption{Confusion matrix for material recognition on a held out test set. Using our learned haptic features, we achieve an accuracy of $42.86\%$.}
  \vspace{-1.5em}
\label{fig:confusion_matrix}
\end{figure}

The features learned from our auto-encoder outperforms sparse coding and hand crafted features for both the deep network and SVM by a significant margin (at least 4.7\%). The feed-forward network also performs at least comparably or slightly better than the SVM for all features. In particular, our haptic features with deep networks improves the traditional method of sparse coding with SVM by $5.8\%$. Furthermore, we show the confusion matrix for material recognition in Fig~\ref{fig:confusion_matrix}. The majority of the error comes from hard objects that are composed of wood/metal/glass being mis-classified as hard plastic. This result demonstrates that our haptic features encode physical properties of the object.

\subsubsection{Grasp Stability Estimation}
The task is to estimate whether the grasp will be successful given tactile signals during grasping. Again, all features are learned from the training set. We train the classifiers on the training set and apply them on our full test set ($580$ trials on $20$ unseen objects). We report the accuracy for binary classification. The results are summarized in Table~\ref{tb:GraspStabilityEstimation}.

\begin{table}[H]
\footnotesize
\centering
\caption{Results of Grasp Stability Estimation}
\vspace{-1em}
\label{tb:GraspStabilityEstimation}
\begin{tabular}{c|cc}
\hline
\multirow{2}{*}{Feature Type} & \multicolumn{2}{c}{Accuracy (\%)} \\
   & Deep Network      & SVM    \\
\hline
Auto-encoder (Ours)                & \textbf{85.92}        & 84.50       \\
Sparse Coding \cite{roberge2016sparsecoding, madry2014st}                    & 81.37       & 80.12     \\
Hand Crafted \cite{spiers2016}                    & 82.54       & 82.66     \\
\hline
\end{tabular}
\vspace{-1.5em}
\end{table}

\begin{figure}[t]
  \begin{center}
    \includegraphics[width = 0.9\linewidth]{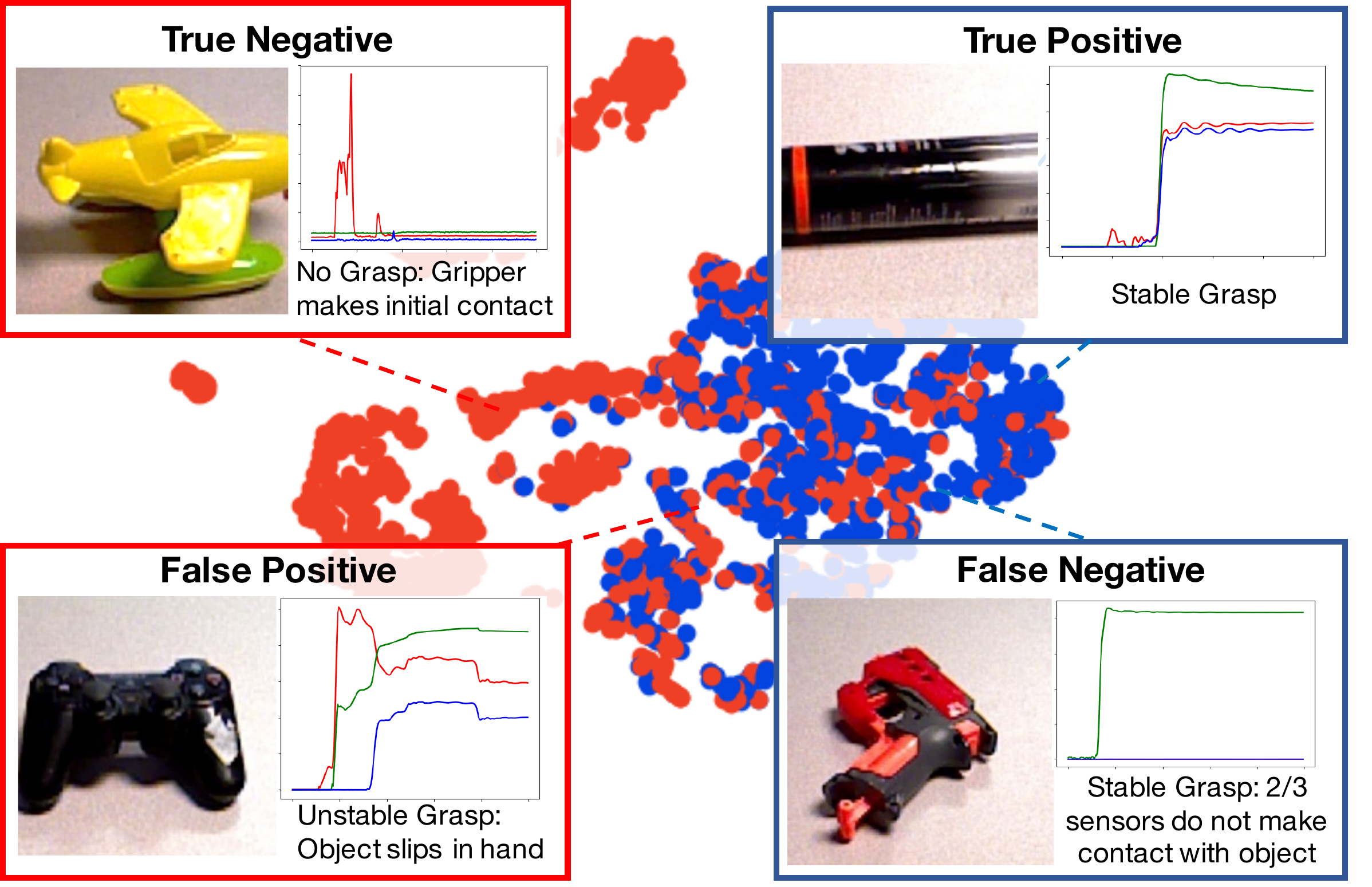}
  \end{center}
  \vspace{-1.6em}
  \caption{Visualization of learned haptic features using t-SNE Embedding. Red and blue dots correspond to failed and successful grasps respectively. We also plot four typical examples for grasp stability estimation.}
\label{fig:tsne}
\vspace{-1.5em}
\end{figure}

The results of grasp stability follow the same trend of material recognition. Our haptic features significantly outperform other features. And the combination of our learned haptic features with deep network achieves the best accuracy. This result suggest that the learned haptic features contains important information for grasping. To better understand our tactile features for grasping, we visualize the t-SNE embedding of the learned features and plot example results of our grasp stability estimation in Fig~\ref{fig:tsne}. We observe that the main failure modes are from that (1) the part of the finger containing the haptic sensor may not come into contact with the object; and (2) the object may slip in the gripper.  

\subsubsection{Remarks} We demonstrate that our learned features are highly effective for two key tactile perception tasks. When compared to other haptic features, our feature learning can  substantially improve the performance. We also show that deep networks are on average better than classical linear SVM with all haptic features. These results provide a strong support to our design of the re-grasping model, i.e.\ the combination of our learned haptic features and deep networks.

\subsection{Tactile Based Grasping}

Our second experiment focuses on the tactile based grasping framework. We first evaluate our core touch based re-grasping model. We then benchmark the full pipeline, and explore incorporating vision based grasping with our re-grasping model.

\subsubsection{Re-grasping Model}
We evaluate our core re-grasping model using the full test set (20 objects). Note in this case, we assume an oracle object location is given: we place each object in eight canonical orientations (N,S,W,E,NE,SE,SW and NW). Moreover, the initial grasp is randomly selected given the object location. We then compare three different settings: a single random re-grasp, multiple random re-grasps, and our re-grasping model. For fair comparison, we set the number of trials for random re-grasps equal to the maximum number of trials of our model. Both random re-grasp and our model are based on our grasp stability estimation. 

\begin{table}[H]
\footnotesize
\centering
\caption{Re-grasping results with oracle object locations}
\vspace{-1em}
\label{tb:VisionTest}
\begin{tabular}{c|c|c|ccc}
\hline
\multirow{2}{*}{\begin{tabular}[c]{@{}c@{}}Object \\ Location\end{tabular}} & \multirow{2}{*}{\begin{tabular}[c]{@{}c@{}}Initial\\ Grasp\end{tabular}} & \multirow{2}{*}{Re-grasp} & \multicolumn{3}{c}{Grasp Accuracy  (\%)} \\
                                                                            &                                                                          &                           & Set A      & Set B     & A+B     \\
\hline
Vision                                                                      & Random                                                                   & -                         & 16.3       & 32.5      & 24.4      \\
Vision                                                                      & Random                                                                   & Random ($\le$4 trials)                & 17.5       & 28.8      & 23.2      \\
Vision                                                                      & Random                                                                   & Ours ($\le$4 trials)                    & \textbf{33.8}       & \textbf{41.3}      & \textbf{37.5}     \\
\hline
\end{tabular}
\vspace{-1.5em}
\end{table}

The results are shown in Table~\ref{tb:VisionTest}. Grasp accuracy on Set B is always higher than Set A. For the full set, the baseline accuracy for chance grasping is $24.4\%$, where the first (and only) grasp is sampled from a random policy with no re-grasping. Interestingly, multiple random re-grasps slightly decreased the accuracy by $1.2\%$. And our re-grasping model get the best accuracy of $37.5\%$. This is $13.1\%$ better than the baseline of multiple random re-grasps. This result demonstrates the effectiveness of our re-grasping module. 


\subsubsection{Grasping without Seeing}
Going beyond re-grasping, we test our full pipeline of tactile based grasping, which includes touch based localization and re-grasping. In this case, we simplify our benchmark by only considering our test set B and use $5$ trials per object. This is primarily limited by the run time of our experiments. 
Our results are show in Table~\ref{tb:TouchTest}. Our pipeline increases the baseline of random grasping by $14\%$ and reaches an accuracy of $40\%$ with only tactile sensing. This is one of the first results for a complete grasping of multiple novel objects using only the sense of touch. 

\begin{table}[H]
\footnotesize
\centering
\caption{Grasping accuracy of our full method. We also present results of combining our re-grasping module with a vision based policy to further improve grasping.}
\vspace{-1em}
\label{tb:TouchTest}
\begin{tabular}{c|c|c|c}
\hline
\begin{tabular}[c]{@{}c@{}}Object \\ Location\end{tabular} & \begin{tabular}[c]{@{}c@{}}Initial \\ Grasp\end{tabular} & Re-grasp & \begin{tabular}[c]{@{}c@{}}Grasp \\ Accuracy (\% on Set B)\end{tabular} \\ \hline
Touch           & Random        & -        & 26.0                   \\
Touch           & Random        & Ours ($\le$4 trials)    & \textbf{40.0}                   \\ \hline
Vision          & Vision        & -        & 51.3                   \\ 
Vision          & Vision        & Ours ($\le$4 trials)    & \textbf{61.9}                   \\ \hline
\end{tabular}
\vspace{-1.5em}
\end{table}


\subsubsection{Visio-Haptic Grasping}
Our last experiment combines the proposed re-grasping model with a vision based policy from ~\cite{murali2017cassl}. The results are show in Table~\ref{tb:TouchTest}. Our framework can further benefit from a good initial grasp (+$11.3\%$). And more importantly, combining vision based grasping with our tactile based re-grasping can largely improve the accuracy by $10.6\%$. These results provide a strong evidence for the need of combining visual and tactile sensing for robust grasping. Through this experiment, we also shows the flexibility of our re-grasping model, which can be readily plugged into existing grasp planning methods.

\section{Conclusions}

In this paper, we demonstrate one of the first attempts of learning to grasp novel objects using only tactile sensing and without prior knowledge about the object. The core of our method lies in the combination of a) a simple method of touch based localization b) unsupervised learning of rich tactile features and c) a learning based method for re-grasping using haptic feedback. First, we created a large-scale dataset for visio-haptic grasping to evaluate our method and to facilitate future research. With this dataset, we used a auto-encoder to learn rich features from raw tactile signals. These features proved effective for both passive tasks like material recognition and active tasks like re-grasping, and displayed an improvement of around $4$-$9$\% over prior methods. Finally, we show that our novel re-grasping model can progressively improve the grasping, leading to significantly higher success rate even from a noisy initial grasp. Our method achieved a grasping accuracy of $40.0\%$ using only tactile sensing for both localization and grasping. We also demonstrate that this re-grasping model can be combined with existing vision based grasping to further improve the accuracy by about 10\%. We hope that our method together with our dataset could provide valuable insights for solving the challenging problem of autonomous grasping.


\section{Future Work}

Our current method is limited in the sense that re-grasping has to start from a random initial grasp, which is far from optimal. Looking forward, tactile exploration could be used to build a representation of object shape (e.g., Gaussian Process Implicit Surfaces) followed by grasp planning~\cite{mahler2015gpgpisopt}. Also, the major failure mode with our current hardware setup is one of partial observability - the regions of the robot's finger not covered by the sensor might come in contact and push the object. This in turns affects all stages of our pipeline - from feature learning, localization, grasp stability estimation to re-grasping. This could be mitigated by using novel skin/contact sensors and wrist force-torque sensors alongside incidental contact algorithms~\cite{tapo2013}. Furthermore, instead of adding symmetric Gaussian noise in the motion model of the particle filter, we can bias the model in the direction of the detected contact force. Finally, a joint learning of localization and re-grasping with reinforcement learning is interesting to explore. Staged learning or policy iteration on the learned policy would greatly improve its performance as in prior work \cite{murali2017cassl, levine2016learning, pinto2016supersizing}.

~\newline
\small
\noindent \textbf{ACKNOWLEDGEMENTS} This work was supported by ONR MURI N000141612007, NSF IIS-1320083 and Google Focused Award. Abhinav Gupta was supported in part by Sloan Research Fellowship and Adithya was partly supported by a Uber Fellowship. The authors would also like to thank Brian Okorn, Nick Rhinehart, Ankit Bhatia, Reuben Aronson, Lerrel Pinto, Tess Hellebrekers and Suren Jayasuriya for discussions.


\small
\bibliographystyle{IEEEtran}
\bibliography{IEEEabrv,references}
\end{document}